\documentclass[runningheads]{llncs}

\usepackage{eccv}

\usepackage{eccvabbrv}

\usepackage{graphicx}
\usepackage{booktabs}
\usepackage{enumitem}

\usepackage{xcolor}
\usepackage{wrapfig}
\usepackage{subcaption}
\usepackage{multirow}
\usepackage{multicol}
\usepackage{indentfirst}
\usepackage{mathtools}
\DeclareMathOperator*{\argmin}{argmin}
\DeclareMathOperator*{\argmax}{argmax}
\newcommand\blfootnote[1]{%
  \begingroup
  \renewcommand\thefootnote{}\footnote{#1}%
  \addtocounter{footnote}{-1}%
  \endgroup
}

\usepackage[accsupp]{axessibility}

\usepackage[pagebackref,breaklinks,colorlinks,citecolor=eccvblue]{hyperref}

\begin{document}

\title{Semantic Prompting with Image-Token for Continual Learning} 

\author{Jisu Han\inst{1} \and
Jaemin Na\inst{2}$^{\ast}$ \and
Wonjun Hwang\inst{1}$^{\ast}$}

\authorrunning{J.~Han et al.}
\titlerunning{I-Prompt}

\institute{Ajou University \and
Tech. Innovation Group, KT\\
\email{jisu3709@ajou.ac.kr; jaemin.na@kt.com; wjhwang@ajou.ac.kr }}

\maketitle

\begin{abstract}
Continual learning aims to refine model parameters for new tasks while retaining knowledge from previous tasks. Recently, prompt-based learning has emerged to leverage pre-trained models to be prompted to learn subsequent tasks without the reliance on the rehearsal buffer. Although this approach has demonstrated outstanding results, existing methods depend on preceding task-selection process to choose appropriate prompts. However, imperfectness in task-selection may lead to negative impacts on the performance particularly in the scenarios where the number of tasks is large or task distributions are imbalanced. To address this issue, we introduce I-Prompt, a task-agnostic approach focuses on the visual semantic information of image tokens to eliminate task prediction. Our method consists of semantic prompt matching, which determines prompts based on similarities between tokens, and image token-level prompting, which applies prompts directly to image tokens in the intermediate layers.
Consequently, our method achieves competitive performance on four benchmarks while significantly reducing training time compared to state-of-the-art methods. Moreover, we demonstrate the superiority of our method across various scenarios through extensive experiments. \blfootnote{$^{\ast}$ Corresponding authors.}
\end{abstract}

\section{Introduction}
Continual learning is an adaptive approach to training deep neural networks, enabling them to adapt and evolve as they encounter new data streams over time. In contrast to traditional paradigms that generally train on static dataset, continual learning focuses on the ability of networks to continuously learn from non-stationary distributions. This approach is crucial in real-world applications where the nature of tasks can dynamically change or expand. The core challenge in continual learning is to equip the networks with the capability to integrate new knowledge while retaining previous knowledge, alleviating catastrophic forgetting~\cite{mccloskey1989catastrophic,goodfellow2013empirical}.

The rehearsal-based approach~\cite{rebuffi2017icarl,bic,er}, widely used in continual learning, aims to mitigate the loss of prior knowledge by periodically retraining the neural network on a subset of previous data using a memory buffer. Although the rehearsal-based methods have demonstrated impressive results in addressing catastrophic forgetting problem, they also raise concerns regarding data privacy and the requirement for additional memory buffers. Consequently, there is a growing demand for the development of a more efficient, rehearsal-free approach~\cite{pass,il2a,ssre,fetril} that can achieve comparable or better performance than current rehearsal-based methods.

\begin{figure}[t]
\begin{center}
\includegraphics[width=1.0\columnwidth]{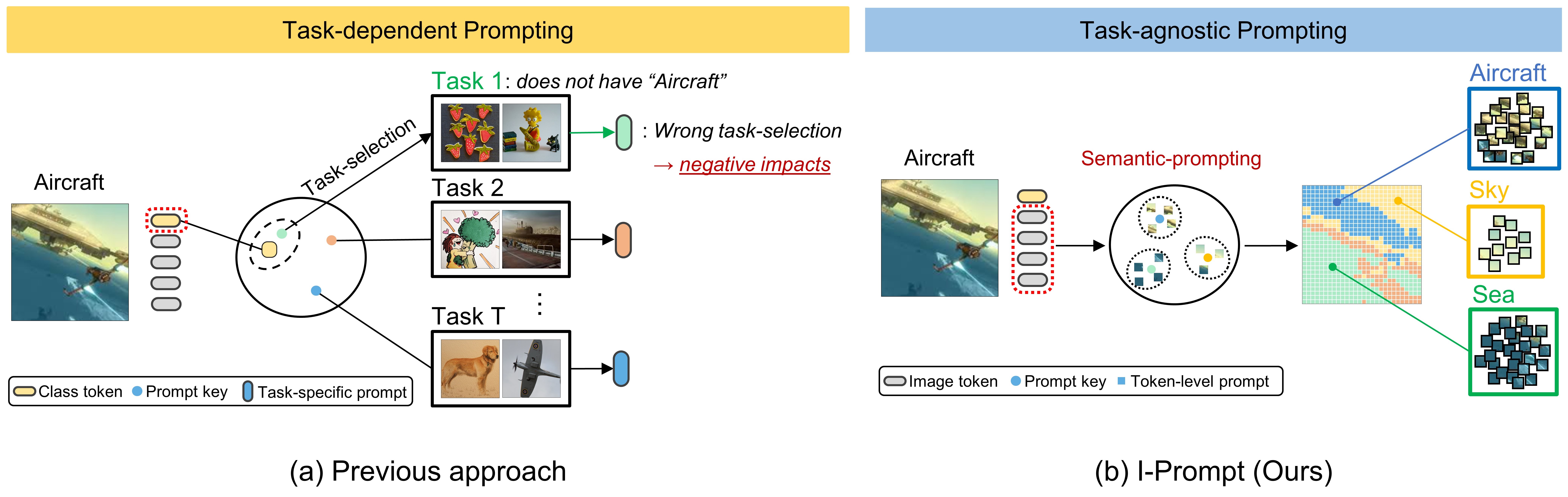}
\end{center}
    \caption{\textbf{Overview of the prompt-based continual learning approaches.} (a) Previous approaches have selected prompts based on the output class token of the pre-trained model and their similarity to task-specific prompts, accompanied by a task selection process.
    If an aircraft image is properly allocated to task T, which involves aircraft, accurate inference can be expected. However, if it is assigned to task 1, it leads to forgetting due to the inconsistency between training and inference.
    (b) Our approach eliminates these erroneous task-selection process and focuses on semantic information within image itself to assign prompts that are relevant to the image. We exploit the information of relationships between image-tokens through the representational capability of the pre-trained model.}
\label{fig:main_1}
\end{figure}

Recently, prompt-based methods~\cite{l2p,dualprompt,coda} have outperformed the rehearsal-based methods, utilizing only a few model parameters without any memory buffers. These methods enhance the performance by training the carefully designed prompts, while keeping the remaining parameters frozen. The latest methods~\cite{l2p} construct a prompt pool and match the output of a pre-trained model as a query to select appropriate prompts from the pool. Building upon this foundation, recent works~\cite{dualprompt,coda} improve efficacy by optimizing the positions of the prompts within the model and refining key-query mechanism. These methods have achieved remarkable success;
however, these methods focus on the task and design task-specific prompts, which involves a task prediction process to select the trained prompts for each task. Therefore, in contrast to the learning process, the selection of prompts that are not trained for the input class due to incorrect task prediction in the inference process where the task ID is unknown leads to forgetting. In soft-selection using weighted combination, forgetting is also caused by the inconsistency of the training and inference process. In particular, as the task prediction becomes more difficult, such as the number of classes per task is imbalanced or the boundary between tasks is blurred, the performance decreases due to the wrong prompt selection and it is difficult to be adaptable to various scenarios.

In this work, we introduce I-Prompt, an image token-based semantic prompting method that exploits the inherent semantic information of image classes. As depicted in Figure~\ref{fig:main_1}, we mitigate the risk of selecting the wrong task by eliminating the traditional task selection process. Instead, we prioritize the use of image tokens to effectively harness the rich semantic information contained within the image itself. 
Our work stems from empirical studies on vision transformer~\cite{vit}, which have demonstrated the effectiveness of exploiting the attention structure in transformer layers for clustering similar tokens~\cite{tokenpooling}. Similarly, we are inspired by previous work~\cite{tokenmerging} that had showed token similarity can be efficiently computed using only self-attention key, which reduces computational costs. Building on these insights, we introduce a prompting method that not only utilizes token similarities within images, but also leverages information about class-specific visual characteristics, offering a more efficient approach.

In continual learning, the prompt query is designed to be applied differently depending on the characteristics of the data to overcome forgetting. In traditional methods, the role of class tokens is to predict the task to which they belong, leveraging the zero-shot classification ability of pre-trained models. Our motivation lies in employing the classification ability of the tokens in the attention structure. This allows us to exploit visual features in the intermediate process, rather than predicting the task by replacing the class token. Moreover, by replacing the task prediction process, we not only reduce the task dependency, but also reduce the additional training cost of the prompt selection process. 
The traditional method of using the class token as a query for the prompt requires an additional forward pass in the prompt selection process to use the final output of the model, resulting in the inefficiency of performing two forward passes. In contrast, our method simplifies this process by selecting the prompts within the transform layers effectively eliminating the need for extra forward pass. Consequently, our approach enables training and inference only with a single forward pass since the prompt selection and prediction process is conducted simultaneously.

We conducted extensive ablation studies for a detailed analysis of the proposed methods. In particular, we achieve competitive performance to the recent state-of-the-art methods in standard task-balanced benchmarks such as CIFAR-100~\cite{cifar}, CUB-200~\cite{cub}, and ImageNet-R/A~\cite{inr, ina}. Furthermore, we demonstrate the effectiveness of our task-agnostic approach in task-imbalanced scenarios, particularly on the ImageNet-R and CIFAR-100 benchmarks. Overall, we make the following contributions:

\begin{itemize}[leftmargin=*,topsep=5pt]
    \item We propose I-Prompt, a novel approach for prompt-based continual learning, which utilizes image token-based semantic prompting to mitigate forgetting.
    \item We address the task dependency issue for the first time in prompt-based methods. Moreover, we achieve a task agnostic prompting method by leveraging visual information from the image itself rather than task information.
    \item  By integrating prompt selection and prediction in a single forward pass, it provides improved efficiency in both the training and inference processes.
    \item  Our approach surpasses the performance compared with the state-of-the-art methods on four benchmarks and is further validated in comprehensive ablation studies.
\end{itemize}

\section{Related work}
\subsection{Continual Learning}
Continual learning aims to enhance performance by acquiring new tasks while minimizing the forgetting of knowledge from previous tasks. Representative methods for solving the forgetting problem in continual learning include regularization-based method, rehearsal-based method, and dynamic architecture method.
Regularization-based methods~\cite{ewc, rwalk, mas, si} determine the importance of model parameters for previous tasks and then apply strong regularization to these important parameters while using less important parameters for learning new tasks. It mitigates forgetting by making the change in loss to the previous task small, while learning new tasks with less important parameters.
This approach offers the benefit of reduced memory requirements for continuous tasks, however it faces the challenge of diminished performance, attributed to updating model parameters without direct access to the data for previous tasks.
Meanwhile, rehearsal-based methods~\cite{rebuffi2017icarl, bic, liu2020mnemonics, prabhu2020gdumb} have achieved high performance by mitigating the forgetting problem through limited-size of memory buffers for the previous tasks. However, they pose additional memory requirements and raise privacy and security concerns due to the storage of past task data. On the other hand, dynamic architecture methods~\cite{pnn,packnet,wang2022foster} freeze models from previous tasks and add sub-networks for the new tasks.
This method effectively avoids forgetting problem but results in a linear increase in model parameters with each new task potentially leading to less manageable models in scenarios with various tasks.

\subsection{Prompt-based Continual Learning}
A method for adapting to downstream tasks without updating the model has been proposed in the field of Natural Language Processing (NLP), focusing on finetuning the large language models~\cite{prompt_tuning, prefix_tuning} using learnable prompts. This success in NLP tasks is extended to vision tasks that require parameter-efficient finetuning in recent studies~\cite{vpt,vqt,vptgen}.
In continual learning, prompt-based methods~\cite{l2p,dualprompt,coda} are included in dynamic architecture method in that there is an additional parameter called a prompt in addition to the model parameters.
L2P~\cite{l2p}, the first study to employ prompts in continual learning, achieves meaningful results by selecting prompts through query-key matching in the query function and learning only the prompts, using them as additional inputs to the model. DualPrompt~\cite{dualprompt} introduces a task-invariant and a task-specific prompts for complementary learning. Furthermore, CODA-Prompt~\cite{coda}, points out the limitation in query-key matching, where the gradient does not flow end-to-end, and addresses this by enhancing learnability through end-to-end training of a prompt directly from the classification loss. Furthermore, language-guided prompt-based approaches~\cite{languageinduce,vlmnotforget} have been studied, but they require an additional memory usage for text encoder.

\subsection{Token Similarity in Transformer}
Research on efficient transformers~\cite{etransformer_adaptive,etransformer_adavit,etransformer_power} is underway to reduce redundant calculations and enable faster calculations by downsampling using similarities between tokens. Token pooling~\cite{tokenpooling} shows that the attention layer of the vision transformer generates overlapping tokens and proposes a method for selecting and pooling similar tokens via clustering from features. Furthermore, Token merging~\cite{tokenmerging} focuses on the self-attention mechanism of the transformer and explains that the key, calculated by cosine similarity for the query, inherently contains token information.
Our method is motivated by findings in the literature on efficient transformers that cluster and token similarity can be calculated through the attention structure in a transformer. We improve efficiency by obtaining the query on which the prompts are selected from inside the transformer layer, instead of using output of a pre-trained transformer encoder.

\section{Method}
\subsection{Preliminary}
\noindent\textbf{Continual learning protocol.}
Continual Learning sets up a learning scenario for sequential tasks $\mathcal{T}=\{1,2,3,...,T\}$, where $T$ denotes the number of total tasks. 
Model consists of a feature extractor and a classifier, each of which is a parametric model with $\theta$ and $\phi$ as parameters. Model prediction $\hat y = f_\phi \cdot f_\theta(x)$, where $f_\theta$ and $f_\phi$ are the feature extractor and classifier, respectively.
We aim to achieve high performance on the test data $D^{1:t}=\{x^{1:t},y^{1:t}\}$ of all previously trained tasks while training on the current task data $D^{t}=\{x^{t},y^{t}\}$ for sequential tasks, where $x$ and $y$ denote image and label, respectively.
Following existing continual learning studies~\cite{l2p,dualprompt,coda}, we focus on class incremental learning scenario. In the class incremental learning scenario, it is assumed that classes for different tasks do not overlap ($y^{t} \cap y^{1:t-1}=\varnothing$), and there is no information about which task is in the test process.

\noindent\textbf{Vision transformer.}
Vision Transformer (ViT) tokenizes the input image, divides it into patches, and then goes through the embedding layer and positional encoding as input to the transformer layer. Therefore, where the feature for $l$-th layer is $h_l$, the initial input to the transformer layer is expressed as $h_0 = [\mathrm{CLS;IMG_1,IMG_2, ...,IMG_p}]$, where p is number of patches. The inside of the transformer layer consists of a Multi-Layer Perceptron ($\mathrm{MLP}$), LayerNorm ($\mathrm{LN}$), Multi-Head Self-Attention ($\mathrm{MHSA}$) and residual connection. The process for each layer is conducted as follows:
\begin{align}
\begin{split}
    &h_{l+1} = \mathrm{MLP}( \mathrm{LN}(z_{l+1}) + z_{l+ 1} ),\\
    &where \  z_{l+1} = \mathrm{MHSA}(\mathrm{LN}(h_l) ) + h_l.
\end{split}
\end{align}

The overall ViT process consists of three steps. First, the input image goes through tokenization and positional encoding for location information. After the class token is combined with the image token obtained, the encoder output is determined through several transformer layers. Finally, the class token from the encoder output is used as input of the classifier to compute the final prediction.

\noindent\textbf{Traditional prompt matching.}
Prompt-based continual learning uses pre-trained ViT as the feature extractor. Since pre-trained models produce consistent output for the same input, existing studies use this process as a query function to select a task-wise prompt. Query-key matching method~\cite{l2p,coda} selects the prompt with the maximum similarity between the query and the prompt key obtained by query function.
\begin{align}
K_s = \argmax\limits_{i \ \in \ \tau} \ \gamma(q(x),k_{i}),
\end{align}
where $K_s$ is selected task-specific prompt key and is an element of the prompt key set $K=\{k_1,k_2, ..., k_{T}\}$, $\tau=\{1,2,3,...,t\}$ is a subset of $\mathcal{T}$ and is the set of tasks up to the current task $t$. Query function $q(x) = f_\theta(x)[\mathrm{CLS}]$ is the class token of the ViT encoder output, and $\gamma(\cdot, \cdot)$ denotes cosine similarity. On the other hand, attention-based method~\cite{coda} performs soft selection using attention rather than hard selection through similarity between queries and keys, and match prompts by their weighted combination. Then, the attention-based prompt $\mathbf{P}_{a}$ is defined as follows:
\begin{align}
\mathbf{P}_{a} = \sum_{i \in \tau} \gamma(q(x)\odot A_{i},k_{i}) \ P_{i},
\end{align}
where prompt $P_{i}$ is element of prompt pool $P=\{P_1,P_2, ..., P_{T}\}$, $A_{i}$ indicates attention, and $\odot$ is element-wise product.
\begin{figure}[t]
\begin{center}
\includegraphics[width=.8\columnwidth]{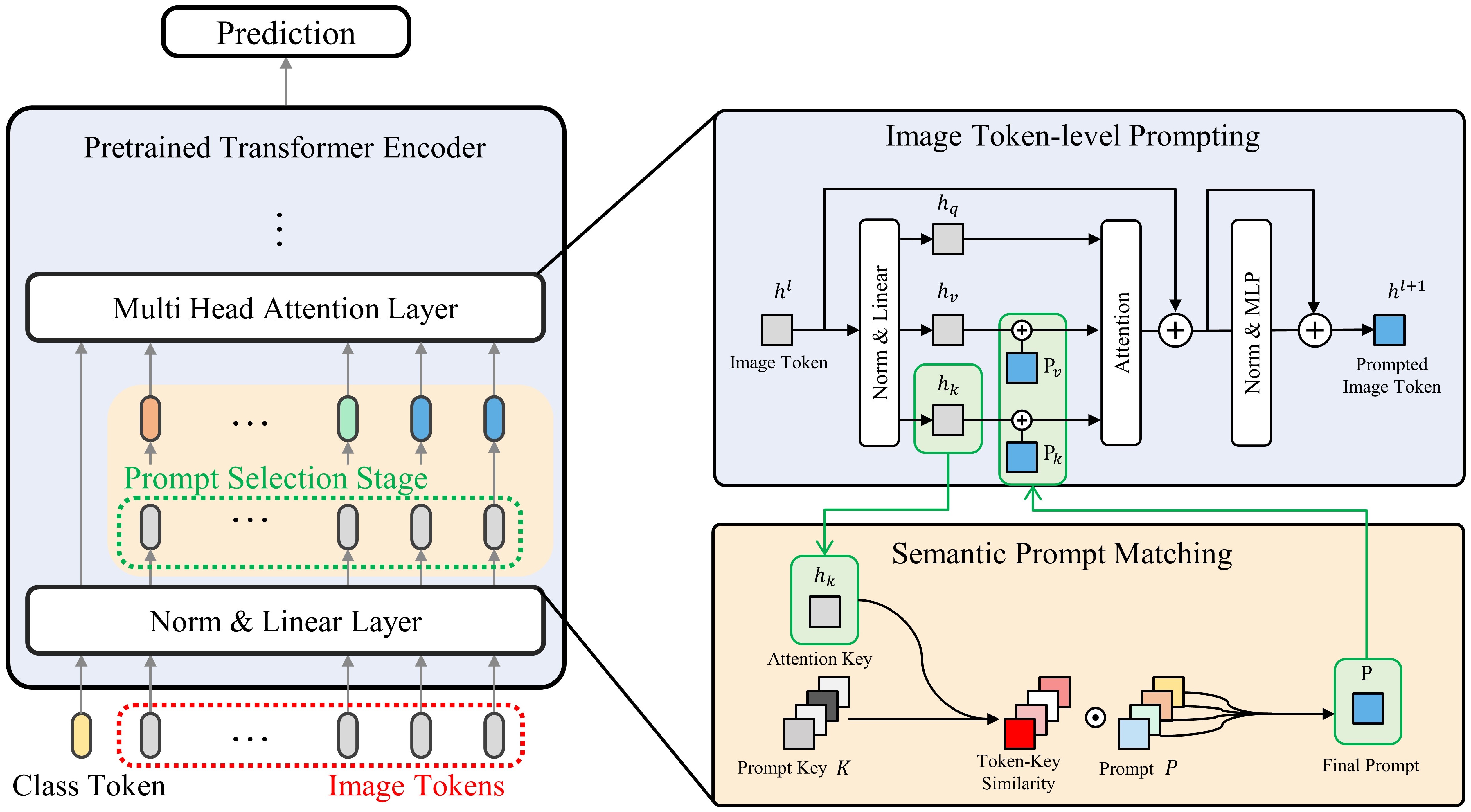}
\end{center}
   \caption{\textbf{Schematic illustration of I-Prompt.}  We describe the details in internal structure of the transformer layer and its interaction with the prompt pool. \textbf{Above:} Transformer layer internal process. The attention key $h_k$ is passed to the prompt pool. Each prompt selected from the prompt pool is added to the transformer's attention key and value, and only the prompt is trained to adapt to new tasks.
   \textbf{Below:} The process of matching prompts in the prompt pool. The similarity between the input attention key from the transformer layer and the prompt key is calculated, and the final prompt is determined by the element-wise product of the calculated similarity and the prompt.
   }
\label{fig:main_2}
\end{figure}

\subsection{Semantic Prompting with Image-token}\label{method}
We aim to develop a task-agnostic method for prompt-based continual learning that eliminates the task prediction process. To achieve this, we select the prompts by focusing on image tokens in the internal structure of the transformer layer, rather than selecting task-wise prompts through a query function.

\noindent\textbf{Semantic prompt matching.}
The self-attention structure in Transformer replicates the query-key-value through a linear layer, and attention to the value is determined based on the cosine similarity between the query and key. In this process, the key in the self-attention can be used as a judge that containing semantic information about the token, which we use as a prompt query for the input token. In order to ensure that similar prompts are assigned to visually similar tokens, the similarity of the prompt keys is calculated based on the self-attention key and used as a weight for each prompt. The calculated weight of the prompt is multiplied by the prompt, and the final prompt for each image token is determined by their summation. Our final prompt $\mathbf{P}$ is defined as follows:
\begin{align}
&\mathbf{P} = \sum_{i \in \tau} \gamma(h_k, k_{i}) \odot P_{i},
\end{align}
where prompt key $k_i$ and prompt $P_i$ are learnable parameters, self-attention key $h_{k} = W_{k}h$ and $W_k$ is weight of self-attention key. The final prompt is split into $P_k$ and $P_v$, which are computed with the key and value of the self attention, respectively.
Semantic prompt matching on similarity between tokens selects prompts through visual representation and thus achieves task-agnostic prompting by focusing on class classification rather than task prediction.

We counteract catastrophic forgetting by fixing previously learned prompts, learning new prompts, and then boosting them. In continual learning, fixing parameters relative to previous learning prompts is the simplest and effective method to deal with catastrophic forgetting~\cite{sprompt,razdaibiedina2023progressive,coda}. However, taking the prompts for all tasks as input leads to a linearly increasing number of input prompts depending on the task. Additionally, considering only the current task fails to take into account class relationships due to the isolation between tasks.
Hence, we propose a more robust prompt $\bar{\mathbf{P}}$ inspired by the boosting algorithm~\cite{wang2022foster}.
\begin{align}
&\bar{\mathbf{P}} = cos(h_k, k_{t}) \odot P_{t} + \sum_{i=1}^{t-1} cos(h_k, \tilde{k}_{i}) \odot \tilde{P}_{i}.
\end{align}
Note that $\tilde{P}_{i}$ and $\tilde{k}_{i}$ are fixed parameters and $P_{t}$ and $k_{t}$ are learnable parameters. Our method achieves a balance between stability and plasticity by carefully learning prompts by fixing the parameters for the previous task and merging the residual for the newly learned task.

\noindent\textbf{Image token-level prompting.}
Prompt tuning methods are divided into prompt-tuning~\cite{prompt_tuning} and prefix-tuning~\cite{prefix_tuning} for input depending on the application location of the prompt.
\begin{flalign}
&\mathrm{Prompt} : h_0 = [\mathrm{CLS;IMG_1,IMG_2, ...,IMG_p};\mathbf{P_s}], \\
&\mathrm{Prefix} : z_{l+1} = \mathrm{MHSA}( \mathrm{LN}([h_l; \mathbf{P_s}]) ) + h_l,
\end{flalign}
where $\mathbf{P_s}$ is the selected prompt, and is determined from the prompt pool. Previous continual learning methods for assigning batch or instance-level prompts are based on these two methods. In contrast, in our method of applying token-level prompts, concatenating prompts on input is inefficient.
For computational efficiency and to take advantage of the prompts selected at the image token level, we adopt a method that directly adds the prompts to the image token:
\begin{align}
&\mathrm{I {\text -} Prompt} : z_{l+1} = \mathrm{MHSA}( \mathrm{LN}([h_l \oplus \bar{\mathbf{P}}]) ) + h_l,
\end{align}
where $\oplus$ denotes element-wise sum. We apply our I-Prompts to the image tokens through a relatively lightweight sum operation, without any dimensional expansion for multiplicative operations. This allows for efficient computation despite our allocation of token-level prompts.
Our method improves classification performance by utilizing image tokens for the final output, since we are training prompts that directly change the image token. The final logit $\hat h$ leveraging the image token is defined as follows:
\begin{align}
\begin{split}
&\hat h = f_{\theta,P,K}(x)[\mathrm{CLS}] + \sum_{i=1}^{p} S(x) \cdot f_{\theta,P,K}(x)[\mathrm{IMG}_i],\\
\end{split}
\end{align}
where the $S(x) = softmax(W_s \gamma(h_k,k))$ is a importance function that determines the importance of each patch in the prediction process through token-key similarity and learnable parameter $W_s$, to influence the final prediction with a large weight for important image tokens.

\noindent\textbf{Objective function.} We optimize the classifier weight $\phi$, prompt $P$ and prompt key $K$ pairs for the fixed model parameters $\tilde\theta$. Finally, our objective function is as follows:
\begin{align}
\argmin\limits_{P,K,\phi} \ \mathcal{L}_{cls} ( f_{\phi} (M \cdot \hat h) ,\ y ),
\end{align}
where $\mathcal{L}_{cls}$ is cross-entropy loss for classification and $M$ is a logit mask, which replaces the logits for classes not included in the training data with negative infinity. Note that the logit mask~\cite{l2p,dualprompt,coda} helps prevent forgetting about previously learned classes by stopping the gradient flow of the classifier for classes that are not training.

\section{Experiments}

\noindent\textbf{Datasets.}
We evaluate our method on continual learning benchmarks such as CIFAR-100~\cite{cifar}, CUB-200~\cite{cub}, renditions of ImageNet~\cite{imagenet}, specifically ImageNet-R/A~\cite{inr,ina}.

\begin{itemize}

\item\textbf{CIFAR-100} is a widely used benchmark for continual learning and contains 100 classes. The train set and test set are split into 50,000 and 10,000 images respectively and consist of the same number of data for all classes.

\item\textbf{ImageNet-R} includes images from various domains to be closer to the real problem. It contains 24,000 training images and 6,000 images for test set. It contains 200 classes.

\item\textbf{ImageNet-A} consists of naturally occurring examples that are incorrectly predicted by models pre-trained with ImageNet and includes 200 classes. The train set and test set are split into 5,981 and 1,519 images.

\item\textbf{CUB-200} is organized into subcategories of birds and contains 200 classes. The training set consists of 9,430 images, and the test set consists of 2,358 images.

\end{itemize}

\noindent\textbf{Evaluation scenarios.}
As a continual learning scenario, we conduct experiments on task-balanced and task-imbalanced scenarios. Task-balanced scenarios are traditional evaluation protocols where the class is equally divided between each task, while task-imbalanced scenarios are settings where the number of classes per task is not constant, including a scenario where half of the class learns on the initial task, and a random increase scenario where the incremental class changes dynamically.
When the initial learning class is \textit{X} and the increasing number of classes is \textit{Y}, it is expressed as \textit{BX-IncY}~\cite{adam}. For example, an experiment that trains 50 classes as base and increases by 10 classes is denoted as \textit{B50-Inc10}.

\noindent\textbf{Evaluation metrics.}
We report Avg-Acc and Last-Acc as evaluation metrics. Let $A_t$ be the average accuracy for all classes in task $t$. where Avg-Acc is the average of the accuracy on each task ($\frac{1}{T} \sum_{t=1}^T A_t$), and Last-Acc is the accuracy for all classes on the final task ($A_T$).

\noindent\textbf{Implementation details.}
Our whole experiments are based on PILOT~\cite{sun2023pilot} and conducted with NVIDIA RTX A6000. We use the Adam optimizer~\cite{kingma2014adam} and adjust learning rate using a cosine scheduling~\cite{loshchilov2017sgdr}, starting from an initial value of 0.001. Following previous works~\cite{dualprompt,coda}, we apply prompts to layers 1-5. We train our method for 20 epochs on the CIFAR-100 and CUB-200, and extend to 50 epochs for ImageNet-R and ImageNet-A. We employ ViT-B/16 model pretrained on ImageNet.
For rehearsal-based methods, we use 20 images as exemplars for each class, and set a fixed size of exemplars for ImangeNet-A with a minimum number of images per class of less than 20. For a fair comparison, we also use the pre-trained ViT-B/16 model as a backbone for the rehearsal-based method.

\subsection{Comparison with State-of-the-Arts}
In our experiments, we compare our method to both rehearsal-based~\cite{rebuffi2017icarl,bic,wang2022foster} and rehearsal-free prompt-based methods~\cite{l2p,dualprompt,coda}. We also provide two distinct methods: Joint-Training, where the entire classes are learned at once, representing the upper bound of accuracy; and Finetuning, conducted without any regularization, which illustrates the lower bound of the accuracy.

\begin{table*}[t]
\centering
\caption{\textbf{Comparison results (\%) in task-imbalanced scenarios.} The methods under comparison are divided into two categories: rehearsal-based and rehearsal-free approaches, with the best accuracy highlighted in bold.}
\label{tab:main-result1}
\resizebox{\textwidth}{!}{  
\begin{tabular}{l| cccccc| cccccc}
\toprule
\multicolumn{1}{c|}{} & \multicolumn{6}{c|}{ImageNet-R} & \multicolumn{6}{c}{CIFAR-100} \\
\cmidrule(lr){2-7} \cmidrule(lr){8-13}
Method        & \multicolumn{2}{c}{B100-Inc5}   & \multicolumn{2}{c}{B100-Inc10}  & \multicolumn{2}{c|}{B100-Inc20} & \multicolumn{2}{c}{B50-Inc2}   & \multicolumn{2}{c}{B50-Inc5} & \multicolumn{2}{c}{B50-Inc10}  \\
               & Avg-Acc     & Last-Acc       & Avg-Acc     & Last-Acc       & Avg-Acc     & Last-Acc     & Avg-Acc     & Last-Acc      & Avg-Acc     & Last-Acc    & Avg-Acc     & Last-Acc      \\
\midrule
Joint-Training            &   -          & 81.58           &   -          & 81.58           &   -          & 81.58         &   -          & 92.33          &   -          & 92.33        &   -          & 92.33    \\
iCaRL            & 68.48        & 60.35           & 68.60        & 60.33           & 71.90        & 64.62         & 80.46        & 68.87          & 83.85        & 73.63        & 86.53        & 79.71    \\
BiC                          & 73.20        & 68.92           & 75.41        & 71.93           & 76.84        & 74.18         & 81.06        & 73.96          & 86.21        & 80.84        & 88.41        & 84.74    \\
Foster              & 80.45        & 77.17           & 80.13        & 76.55           & 79.88        & 76.60         & 90.83        & 87.88          & 90.67        & 88.56        & 90.47        & 87.80    \\
\midrule[0.5pt]
Finetuning       & 59.63        & 47.37           & 64.08        & 57.07           & 72.49        & 61.67         &  67.86        & 60.21          & 78.61        & 69.06        & 81.14        & 73.39    \\
L2P            & 64.07        & 52.65           & 68.84        & 59.58           & 73.16        & 66.63         &  67.67        & 49.80          & 80.21        & 69.27        & 86.78        & 80.57    \\
DualPrompt      & 62.36        & 54.03           & 66.47        & 59.82           & 70.15        & 65.03         &  68.81        & 52.98          & 81.79        & 73.44        & 86.03        & 80.84    \\
CODA-Prompt      & 71.24        & 64.20           & 75.75        & 70.88           & 77.88        & 73.92         &  74.58        & 61.23          & 85.21        & 78.23        & 90.07        & 85.26    \\
\midrule[0.5pt]\midrule[0.5pt]
I-Prompt (Ours)           & \textbf{73.96}        & \textbf{67.30}           & \textbf{78.01}        & \textbf{73.30}           & \textbf{79.23}        & \textbf{75.82}         & \textbf{75.10}        & \textbf{63.26}          & \textbf{86.66}        & \textbf{80.75}        & \textbf{90.32}        & \textbf{87.09}    \\
\bottomrule
\end{tabular}
}
\end{table*}

\begin{table*}[t]
\centering
\caption{\textbf{Comparison results (\%) in task-balanced scenarios.} The comparison methods are categorized into rehearsal-based and rehearsal-free methods.}
\label{tab:main-result2}
\resizebox{\textwidth}{!}{  
\begin{tabular}{l| cccccc| cccccc}
\toprule
\multicolumn{1}{c|}{} & \multicolumn{6}{c|}{ImageNet-R} & \multicolumn{6}{c}{CIFAR-100} \\
\cmidrule(lr){2-7} \cmidrule(lr){8-13}
Method              & \multicolumn{2}{c}{B0-Inc10}   & \multicolumn{2}{c}{B0-Inc20}  & \multicolumn{2}{c|}{B0-Inc40}     & \multicolumn{2}{c}{B0-Inc5}   & \multicolumn{2}{c}{B0-Inc10} & \multicolumn{2}{c}{B0-Inc20}  \\
                    & Avg-Acc     & Last-Acc       & Avg-Acc     & Last-Acc       & Avg-Acc     & Last-Acc     &  Avg-Acc     & Last-Acc      & Avg-Acc     & Last-Acc    & Avg-Acc     & Last-Acc      \\
\midrule
Joint-Training               &   -          & 81.58           &   -          & 81.58           &   -          & 81.58         &    -          & 92.33          &   -          & 92.33        &   -          & 92.33    \\
iCaRL                & 68.50        & 56.28           & 71.82        & 61.43           & 75.65        & 65.60         &  82.71        & 69.52          & 85.30        & 74.60        & 87.61        & 79.20    \\
BiC                  & 78.49        & 71.57           & 80.21        & 74.85           & 81.05        & 76.17         &  85.21        & 77.30          & 88.52        & 82.72        & 90.71        & 86.37    \\
Foster               & 83.00        & 76.52           & 82.46        & 76.27           & 82.32        & 75.73         &  91.87        & 87.22          & 92.08        & 87.83        & 91.41        & 86.80    \\ 
\midrule[0.5pt]
Finetuning            & 66.75        & 48.23           & 71.35        & 61.40           & 76.24        & 64.78         &  75.83        & 63.86          & 79.50        & 68.01        & 84.96        & 75.65    \\
L2P                  & 75.91        & 70.13           & 78.11        & 72.63           & 78.78        & 74.62         &  84.18        & 77.70          & 89.26        & 84.41        & 90.54        & 85.85    \\
DualPrompt           & 72.52        & 66.00           & 74.94        & 69.13           & 74.51        & 70.05         &  84.88        & 77.39          & 87.39        & 82.38        & 88.10        & 83.46    \\
CODA-Prompt           & 78.65        & 72.18           & 81.44        & 75.08           & 81.46        & 76.72         &  88.03        & 80.66          & 91.45        & 86.19        & 92.52        & 88.40    \\
\midrule[0.5pt]\midrule[0.5pt]
I-Prompt (Ours)     & \textbf{79.74}  & \textbf{73.22}        & \textbf{81.75}        & \textbf{75.73}           & \textbf{81.86}        & \textbf{76.92}         &  \textbf{89.69}        & \textbf{84.62}          & \textbf{91.75}        & \textbf{87.63}        & \textbf{92.69}        & \textbf{88.91}    \\
\bottomrule[1.0pt]
\end{tabular}
}
\end{table*}

\noindent\textbf{Task-imbalanced scenario.}
Table~\ref{tab:main-result1} shows the results for the task-imbalanced scenario where the distribution of classes per task is uneven. In this setup, half of all classes are initially trained, and then the remaining classes are split according to tasks. Overall, our method demonstrates superior performance on both ImageNet-R and CIFAR-100 benchmarks, outperforming other state-of-the-art methods. Specifically, in ImageNet-R, we achieve a significant performance improvement of 2.72\% in average accuracy and 3.30\% in last accuracy compared to the existing best accuracy in B100-Inc5 setting. In CIFAR-100, we obtain performance enhancements of up to 1.45\% and 2.52\% for average and last accuracy, respectively, in B50-Inc5 setting. Performance increases significantly as the total number of tasks increases. This tendency satisfies our objective of mitigating the forgetting problem that occurs as task prediction becomes more difficult.

\begin{table*}[t]
\centering
\caption{\textbf{Comparison results (\%) in task-balanced scenarios.} Comparison methods are divided into rehearsal methods and rehearsal-free methods.}
\label{tab:main-result3}
\resizebox{\textwidth}{!}{  
\begin{tabular}{l| cccccc| cccccc}
\toprule
\multicolumn{1}{c|}{} & \multicolumn{6}{c|}{ImageNet-A} & \multicolumn{6}{c}{CUB-200} \\
\cmidrule(lr){2-7} \cmidrule(lr){8-13}
Method        & \multicolumn{2}{c}{B0-Inc10}   & \multicolumn{2}{c}{B0-Inc20}  & \multicolumn{2}{c|}{B0-Inc40}&  \multicolumn{2}{c}{B0-Inc10}   & \multicolumn{2}{c}{B0-Inc20} & \multicolumn{2}{c}{B0-Inc40}  \\
              & Avg-Acc     & Last-Acc       & Avg-Acc     & Last-Acc       & Avg-Acc     & Last-Acc      & Avg-Acc     & Last-Acc      & Avg-Acc     & Last-Acc    & Avg-Acc     & Last-Acc      \\
\midrule
Joint-Training         &   -          & 56.81           &   -          & 56.81           &   -          & 56.81          &   -          & 87.79          &   -          & 87.79        &   -          & 87.79    \\
iCaRL          & 49.12       & 37.33           & 40.15        & 32.13           & 49.09        & 39.83          & 89.59        & 82.44          & 89.81        & 84.56        & 89.07        & 83.50    \\
BiC            & 50.12        & 38.25           & 47.21        & 38.31           & 50.55        & 40.75          & 88.14        & 83.59          & 89.19        & 83.72        & 89.91        & 85.33    \\
Foster         & 60.28        & 50.30           & 57.55        & 50.63           & 55.97        & 49.64          & 85.26        & 82.70          & 81.48        & 77.65        & 78.28        & 70.53    \\
\midrule[0.5pt]
Finetuning      & 30.91        & 13.63           & 36.94        & 20.80           & 44.91        & 26.27          & 57.66        & 36.34          & 69.80        & 52.12        & 77.52        & 65.31    \\
L2P           & 48.73        & 38.78           & 53.10        & 44.70           & 54.82        & 48.12          & 70.02        & 58.06          & 77.27        & 66.54        & 79.90        & 71.46    \\
DualPrompt     & 55.20        & 43.05           & 57.81        & 47.07           & 59.77        & 50.76          & \textbf{77.52}        & 64.80          & 79.91        & 69.17        & 81.57        & 72.73    \\
CODA-Prompt     & 54.33        & 44.63           & 62.85        & 52.40           & 65.74        & 56.22          & 76.77        & \textbf{66.58}          & 83.39        & 73.20        & 85.76        & 77.78    \\
\midrule[0.5pt]\midrule[0.5pt]
I-Prompt (Ours)          & \textbf{62.28}        & \textbf{50.76}           & \textbf{65.71}        & \textbf{55.83}           & \textbf{66.48}        & \textbf{56.48}          & 77.29        & 66.07          & \textbf{84.25}        & \textbf{74.64}        & \textbf{85.81}        & \textbf{78.67}    \\
\bottomrule
\end{tabular}
 }
\end{table*}

\begin{table}[t]
\centering
\caption{\textbf{Results (\%) on online continual learning scenario.} A higher AUC-Acc indicates that model's ability to consistently maintain high accuracy while adapting to new task over the training. $\dagger$ denotes our reproduced results with their official codes.}
\label{tab:Ablation table1}
\resizebox{0.7\textwidth}{!}{
\begin{tabular}{l|cc|cc}
\toprule
\multirow{2.5}{*}{Method} & \multicolumn{2}{c}{CIFAR-100} & \multicolumn{2}{|c}{ImageNet-R} \\
\cmidrule(lr){2-3} \cmidrule(lr){4-5}
                & AUC-Acc       & Last-Acc      & AUC-Acc        & Last-Acc       \\
\midrule
Finetuning      & $19.71\pm3.39$ & $10.42\pm4.92$ & $7.51\pm3.94$   & $2.29\pm0.85$  \\
Linear Probing  & $49.69\pm6.09$ & $23.07\pm7.33$ & $29.24\pm1.26$  & $16.87\pm3.14$  \\
\midrule[0.5pt]
L2P            & $57.08\pm4.43$ & $41.63\pm12.73$ & $29.65\pm1.63$  & $19.55\pm4.78$  \\
DualPrompt      & $67.07\pm4.16$ & $56.82\pm3.49$ & $40.11\pm1.27$  & $29.24\pm4.63$  \\
MVP             & $\textbf{68.10}\pm\textbf{4.91}$ & $62.59\pm2.38$ & $40.60\pm1.21$  & $31.96\pm3.07$  \\
MVP$^\dagger$   & $67.13\pm5.05$ & $63.10\pm1.61$ & $38.39\pm1.54$  & $31.01\pm3.72$  \\
\midrule[0.5pt]\midrule[0.5pt]
I-Prompt (Ours)            & $67.23\pm5.76$ & $\textbf{63.42}\pm\textbf{1.48}$ & $\textbf{41.08}\pm\textbf{1.54}$  & $\textbf{33.27}\pm\textbf{2.86}$  \\
\bottomrule
\end{tabular}
}
\end{table}

\noindent\textbf{Task-balanced scenario.}
We show the experimental results for the task-balanced scenario in Tables~\ref{tab:main-result2} and~\ref{tab:main-result3}. The task-balancing scenario splits all tasks into an equal number of classes. The task-balanced scenario is the basic experimental setting of the previous work~\cite{l2p,coda,dualprompt}, and there is no task imbalance problem. Our method achieves competitive performance on CIFAR-100, ImageNet-R/A, and CUB-200 compared to prompt-based methods. Especially, for CIFAR-100 B0-Inc5, it achieves 1.66\% and 3.96\% higher performance in terms of average accuracy and last accuracy than before, and in ImageNet-A, which has the largest performance gap, it achieves a high average accuracy of up to 7.08\% difference. Moreover, it achieves notable performance even when compared to rehearsal-based methods, and even achieves better performance on CIFAR-100 B0-Inc20. In the task-balanced scenario, the performance improvement is large compared to the comparison method in an experimental environment with many tasks, which is the same as the tendency of the imbalanced scenario.

\subsection{Ablation Studies}

\noindent\textbf{Online continual learning scenario.}
In Table~\ref{tab:Ablation table1}, we validate our method in Si-Blurry~\cite{mvp} scenario characterized by stochastic blurry task boundary, which introduces a more challenging online continual learning scenario. For this experiment, we adopt AUC-Acc~\cite{koh2022online} as a metric to evaluate the efficacy of the methods.
As a result, we have demonstrated the robustness of our approach, consistently achieving strong AUC-Acc and Last-Acc results for both CIFAR-100 and ImageNet-R. We obtain the best performance in both metrics for ImageNet-R. We also achieve the best Last-Acc in CIFAR-100 while obtaining the second-best performance of AUC-Acc. Remarkably, our method achieves competitive AUC-Acc performance without explicitly addressing the inherent class imbalance problem in this scenario, in contrast to the MVP~\cite{mvp}, which employs alignment techniques for this issue.

\begin{figure}[t]
  \centering
  \begin{minipage}{0.44\linewidth}
          \includegraphics[width=\linewidth]{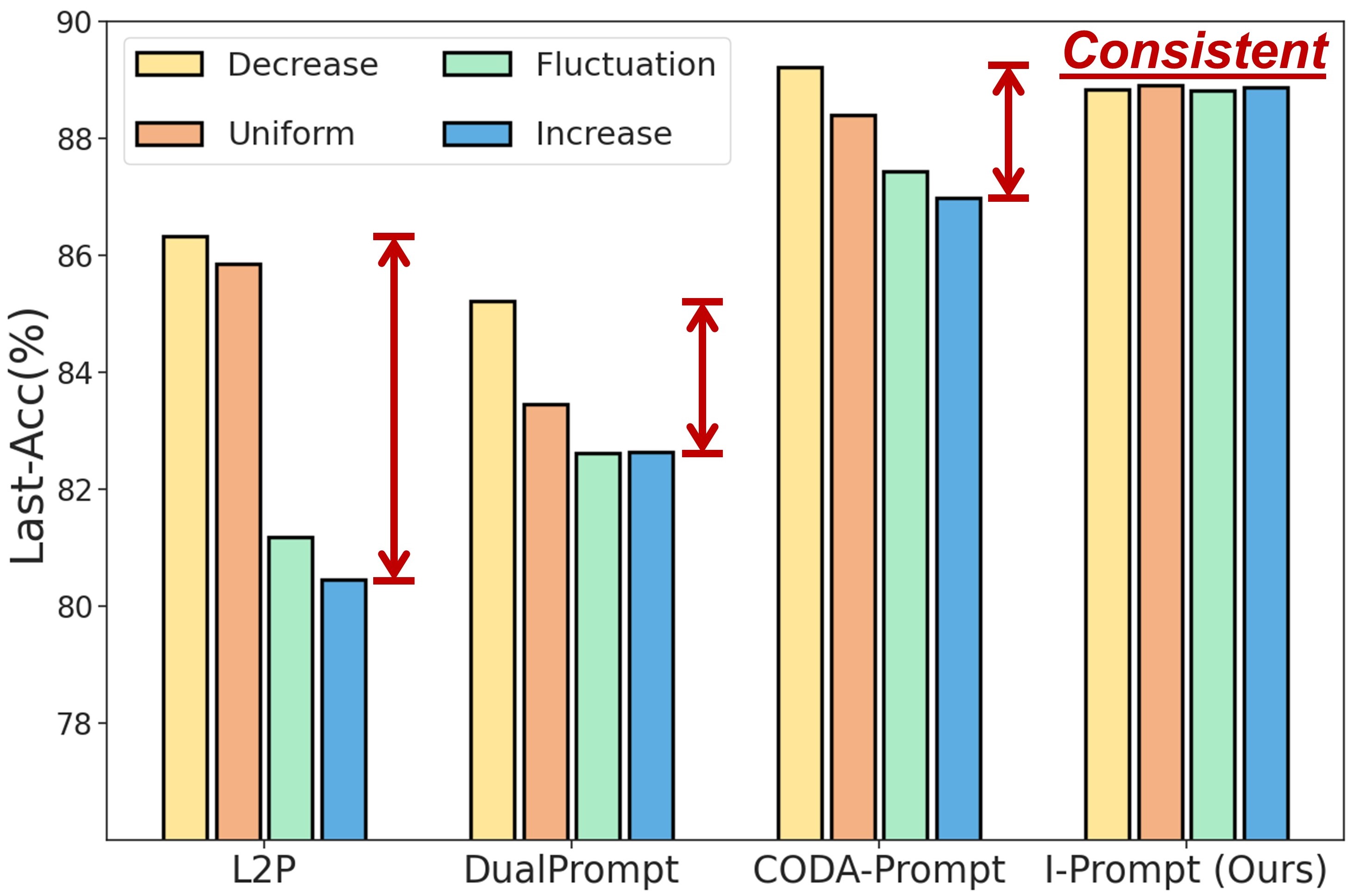}
          \caption{\textbf{Performance on various task distribution.} We report the final accuracy in the uniform setting, representing the task-balanced scenario, and in three distinct task-imbalanced scenarios.}
          \label{fig:abs_1}
  \end{minipage}
  \hspace{0.01\linewidth}
  \begin{minipage}{0.52\linewidth}
          \includegraphics[width=\linewidth]{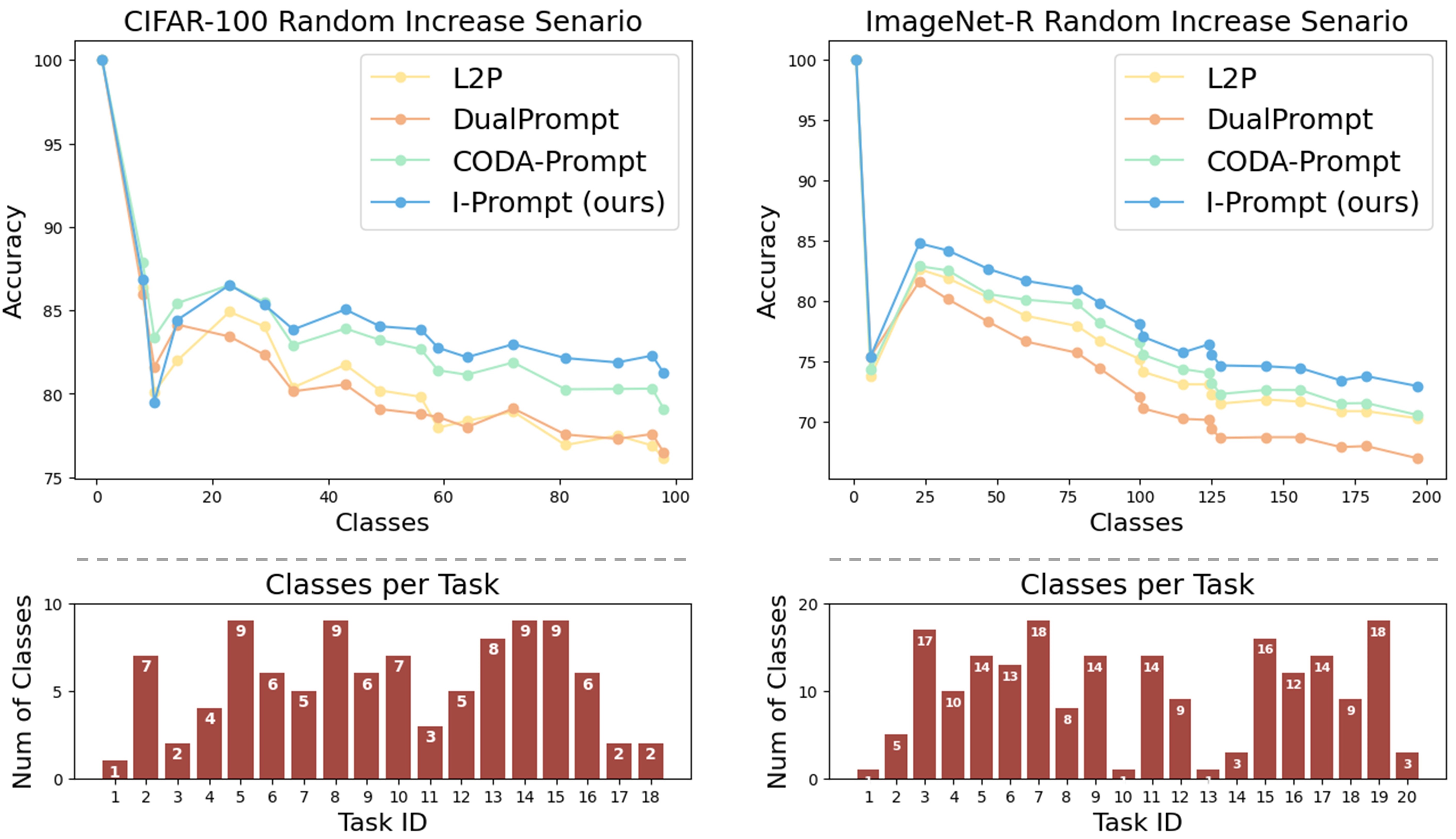}
          \caption{\textbf{Task-wise accuracy in random increase scenario.} We report the performance of the prompt-based method in a random increase scenario. The line plot and bar plot show the average accuracy and the distribution of classes per task, respectively.}
          \label{fig:main_3}
  \end{minipage}
\end{figure}

\noindent\textbf{Various task distribution.}
In Figure~\ref{fig:abs_1}, we present the last accuracy across a variety of task distributions to demonstrate that our task-agnostic method consistently provides robust performance for each distribution. We set up the experiments with four different task distributions: decreasing, uniform, increasing, and fluctuating, according to the number of classes per task. The uniform distribution has a uniform number of classes per task and is equal to B0-Inc20. The decreasing distribution has a decreasing number of classes per task and is equal to B0-Inc(30-5t). The increasing distribution has classes per task increase and is equal to B0-Inc(10+5t). Finally, the fluctuating distribution has increasing and decreasing classes per task and has 10,30,5,40,15 classes for each task. In experiments with uniform and decreasing task distributions, L2P, DualPrompt, and CODA-Prompt demonstrate high performance levels; however, their effectiveness diminishes in scenarios involving increasing or fluctuating distributions. It is highly optimized for the initial task during the prompt selection process, resulting in high performance when the number of initial training classes is large, but due to task dependency, the final accuracy changes significantly as the task changes. On the other hand, the proposed method showed equivalent performance in all experiments. This shows that our method is not task-dependent by fully exploits information about the classes, making it task agnostic.

\noindent\textbf{Random increase scenario.}
As a further task imbalance scenario, we report the task-wise accuracy of recent continual learning methods in Figure~\ref{fig:main_3} for a random increase scenario with dynamically changing incremental steps. Each task is assigned a number of classes generated from the same random seed, and the histogram below shows the distribution of classes per task. In experiments on random increase scenario, we observe a trend that as the number of tasks increases, the performance gap between our method and the comparison methods also widens. This trend highlights the effectiveness of our task-agnostic approach, particularly in addressing the existing challenge where task prediction becomes more difficult with an increasing number of tasks.

\begin{table}[t]
\centering
\caption{\textbf{Efficiency comparison.} We provide the number of training parameters, training and inference times, and accuracy.
}
\label{tab:Ablation table2}
\resizebox{0.7\textwidth}{!}{  
\begin{tabular}{l|c|c|c|c}
\toprule
\multirow{2}{*}{Method}  & Tuning Parameters$\downarrow$ & Training Time$\downarrow$  &Inference Time$\downarrow$ & Avg-Acc$\uparrow$  \\
                         & (learnable/total)                         & (ms/image)               & (ms/image) &                    (\%)             \\
\midrule
L2P             & $0.14\%$        & 8.15  & 2.51   & 89.26\\
DualPrompt      & $0.39\%$        & 7.44  & 2.71   & 87.39\\
CODA-Prompt     & $4.57\%$        & 9.47  & 4.90   & 91.45\\
\midrule[0.5pt]\midrule[0.5pt]
I-Prompt (Ours)              & $1.43\%$     &   \textbf{5.86}  & \textbf{2.50} & \textbf{91.75} \\
\bottomrule
\end{tabular}
}
\end{table}

\begin{figure}[t]
\begin{center}
\includegraphics[width=0.7\linewidth]{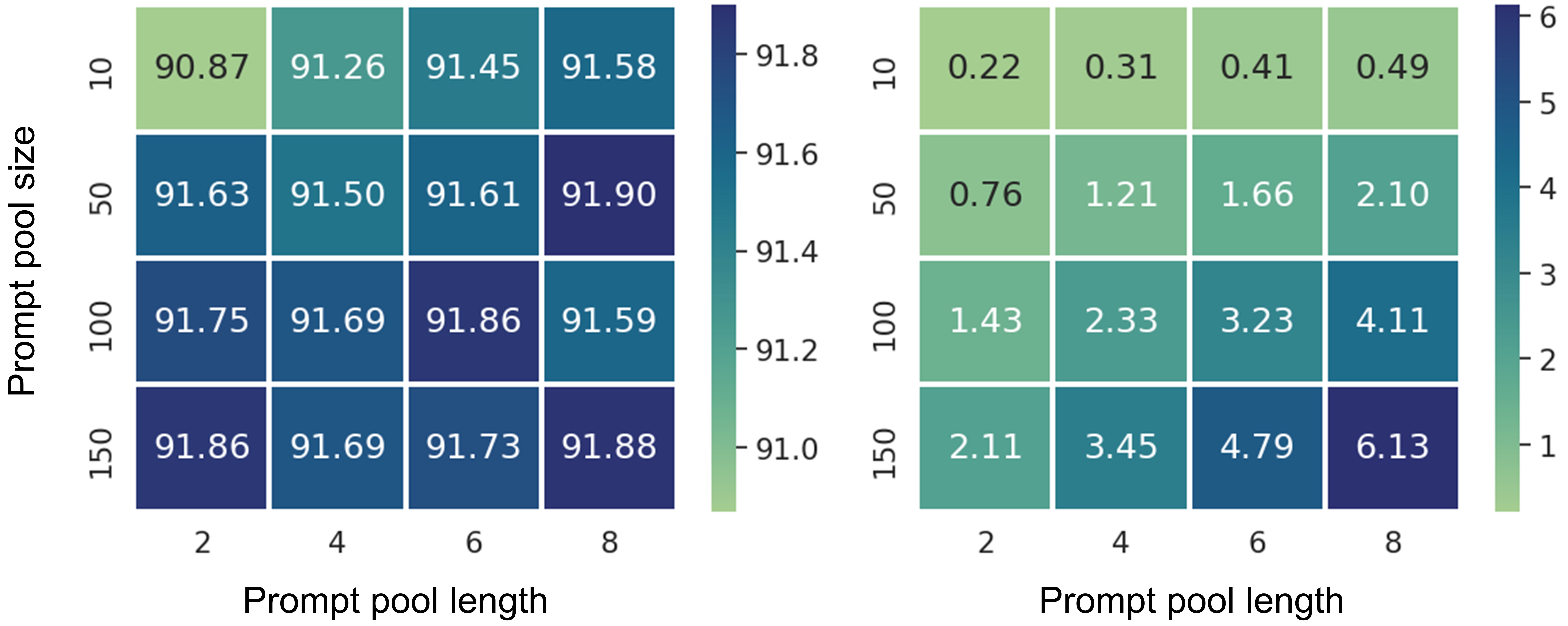}
\end{center}
\caption{\textbf{Hyperparameter analysis.} Result of the grid search for prompt pool size and length, \textbf{Left:} average accuracy $\uparrow$ (\%), \textbf{Right:} tuning parameter ratio $\downarrow$ (\%).}
\label{fig:abs_2}
\end{figure}

\noindent\textbf{Efficiency comparison.}
Table~\ref{tab:Ablation table2} shows the number of training parameters compared to total model parameters as additional memory usage and the learning and inference time as computation cost.
Compared to the previous best performance, our method achieves the best accuracy with only 33\% of the number of trainable parameters. It also achieves a high performance of 90.87\% for 0.22\% training parameters, which is close to the training parameters of L2P and DualPrompt, as shown in Figure~\ref{fig:abs_2}.
For training time and inference time, we achieve a 40-50\% reduction compared to CODA-Prompt, which also achieves the best performance in comparison to L2P and DualPrompt. Our method requires only one forward pass in the training and inference process, while previous methods require an additional forward pass to obtain the query as a selection criterion for the prompt, for a total of two forward passes. This allows our method to train and inference efficiently.

\noindent\textbf{Hyperparameter analysis.}
Figure~\ref{fig:abs_2} shows the average accuracy and number of training parameters by grid search in [10,50,100,150] for prompt pool size and [2,4,6,8] for prompt pool length.
The hyperparameters of our method include the prompt pool size and prompt length. The size of the prompt pool is the total number of prompts, and the prompt length is the number of prompts concatenated to the input.
Since our method adds prompts without concatenate them to inputs, it is redefined by the number of prompts added to each image token.
The prompt is applied separately to the self-attention key and value, thus the prompt pool length is twice the prompt length.
Considering a compromise between average accuracy and the number of training parameters, we empirically chose the prompt pool size and length to be 100 and 2, respectively.

\noindent\textbf{Effects of each component.}
We show the experimental results for each component of the proposed method in Table~\ref{tab:Ablation table4} to investigate the effect of each component on the performance. The proposed method consists of semantic prompt matching and image token-level prompts. The baseline is a structure consisting of only task-specific prompts in dual prompting, with the addition of prefix tuning for the task-specific prompt pool. We compare the performance with image token-level prompting, which directly changes the image token, and with semantic prompt matching, which selects prompts based on internal information rather than task. Both improve performance over the baseline, with the best performance achieved when both are applied.

\begin{table}[t]
\centering
\caption{\textbf{Effects of each component.} We report the performance of each component in our method and overall performance.}
\label{tab:Ablation table4}
\resizebox{0.7\textwidth}{!}{
\begin{tabular}{l cc cc}
\toprule
\multirow{2.5}{*}{Method} & \multicolumn{2}{c}{CIFAR-100} & \multicolumn{2}{c}{ImageNet-R} \\
\cmidrule(lr){2-3} \cmidrule(lr){4-5}
                & Avg-Acc       & Last-Acc      & Avg-Acc        & Last-Acc       \\
\midrule
Baseline                           & 84.06 & 76.95 & 73.75  & 66.93  \\
\midrule
w/ image token-level prompting     & 89.17 & 83.14 & 78.54  & 72.40  \\
w/ semantic prompt matching        & 91.50 & 87.28 & 77.85  & 72.13  \\
w/ both (I-Prompt)                 & 91.75 & 87.63 & 81.75  & 75.73  \\
\bottomrule
\end{tabular}
}
\end{table}

\section{Conclusion}
In this paper, we have presented a novel task-agnostic prompting method, I-Prompt, to address catastrophic forgetting problem in continual learning. Instead of existing task-dependent approaches that rely on task-selection, we focus on the semantic features of the images themselves. As a result, our method not only resolves the negative effects of incorrect task-selection, but also improves training efficiency by compressing the prompt selection process into a single forward pass. Our extensive empirical studies, including both task-balanced and task-imbalanced scenarios, provide in-depth insights into task-agnostic approach in prompt-based continual learning, affirming our method's effectiveness. We hope that our approach could serve as a valuable groundwork for moving towards various continual learning scenarios.

\bibliographystyle{splncs04}
\bibliography{main}
\end{document}